\documentclass{article}

\usepackage[nonatbib, final]{mod_neurips}
\usepackage[numbers]{natbib}
\usepackage[utf8]{inputenc} 
\usepackage[T1]{fontenc}    
\usepackage[hidelinks]{hyperref}       
\usepackage{url}  
\usepackage{booktabs}   
\usepackage{amsfonts}       
\usepackage{nicefrac}       
\usepackage{microtype}      
\usepackage{xcolor}
\usepackage{caption}
\usepackage{microtype}
\usepackage{graphicx}
\usepackage{wrapfig}
\usepackage{subcaption}
\usepackage{amsmath}
\usepackage{xspace}
\usepackage{amsfonts}
\usepackage{amssymb}
\usepackage{amstext}
\usepackage{multirow}
\usepackage{enumitem}
\usepackage{paralist}
\usepackage[ruled, linesnumbered]{algorithm2e}
\usepackage{multicol}
\let\oldnl\nl
\newcommand{\nonl}{\renewcommand{\nl}{\let\nl\oldnl}}

\newcommand{\self}{\ensuremath{\mathsf{SELF}}\xspace}
\newcommand{\mixed}{\ensuremath{\mathsf{MIXED}}\xspace}
\newcommand{\meta}{\ensuremath{\mathsf{META}}\xspace}

\title{Federated Meta-Learning with Fast Convergence and Efficient Communication}

\author{
	Fei Chen$^{*}$\\
	Huawei Noah's Ark Lab \\
	\texttt{chen.f@huawei.com} \\
	\And
	Mi Luo$^{}\thanks{Equal contribution. This work was done when Mi Luo was intern at Huawei Noah's Ark Lab.}$\\
	Huawei Noah's Ark Lab \\
	\texttt{rosemaryluo@outlook.com} \\
	\And
	Zhenhua Dong \\
	Huawei Noah's Ark Lab \\
	\texttt{dongzhenhua@huawei.com} \\
	\And
	Zhenguo Li \\
	Huawei Noah's Ark Lab \\
	\texttt{li.zhenguo@huawei.com} \\
	\And
	Xiuqiang He \\
	Huawei Noah's Ark Lab \\
	\texttt{hexiuqiang1@huawei.com} \\
}

\begin{document}
	
	\maketitle
	
	\begin{abstract}
    Statistical and systematic challenges in collaboratively training machine learning models across distributed networks of mobile devices have been the bottlenecks in the real-world application of federated learning. In this work, we show that meta-learning is a natural choice to handle these issues, and propose a federated meta-learning framework \texttt{FedMeta}, where a parameterized \emph{algorithm} (or \emph{meta-learner}) is shared, instead of a global model in previous approaches. We conduct an extensive empirical evaluation on LEAF datasets and a real-world production dataset, and demonstrate that \texttt{FedMeta} achieves a reduction in required communication cost by 2.82-4.33 times with faster convergence, and an increase in accuracy by 3.23\%-14.84\%  as compared to Federated Averaging (FedAvg) which is a leading optimization algorithm in federated learning. Moreover, \texttt{FedMeta} preserves user privacy since only the parameterized algorithm is transmitted between mobile devices and central servers, and no raw data is collected onto the servers.
    
\end{abstract}

	\section{Introduction}
\label{sec:intro}

The success of deep learning has relied heavily on large amounts of labeled data. In many scenarios, the data is distributed among various clients and is privacy-sensitive, making it unrealistic to collect raw data onto central servers for model training. Meanwhile, as the storage and computational power of mobile devices grow, it is increasingly attractive to move computation, such as the training of machine-learning models, from the cloud to the edge devices. These issues motivate federated learning~\cite{McMahan2016,bonawitz2019towards,konevcny2016federated,li2019federated}, which aims to collaboratively train models by maintaining a shared model on a central server and utilizing all clients' data in a distributed fashion. This setting preserves user privacy without collecting any raw data, while the statistical challenges and systematic challenges become important problems for algorithm design. For statistical challenges, the decentralized data is non-IID, highly personalized and heterogeneous, which leads to a significant reduction in model accuracy~\cite{zhao2018federated}. For systematic challenges, the number of devices is typically order of magnitudes larger than that in the traditional distributed settings. Besides, each device may have significant constraints in terms of storage, computation, and communication capacities. To tackle the two challenges, ~\cite{McMahan2016} proposed the Federated Averaging (FedAvg) algorithm, which can flexibly determine the number of epochs and batch size for local training with SGD, so as to achieve high accuracy as well as trade-off between computation and communication cost.

\begin{figure*}  
\captionsetup{font=small}
    \centering
    \includegraphics[width=0.75\textwidth]{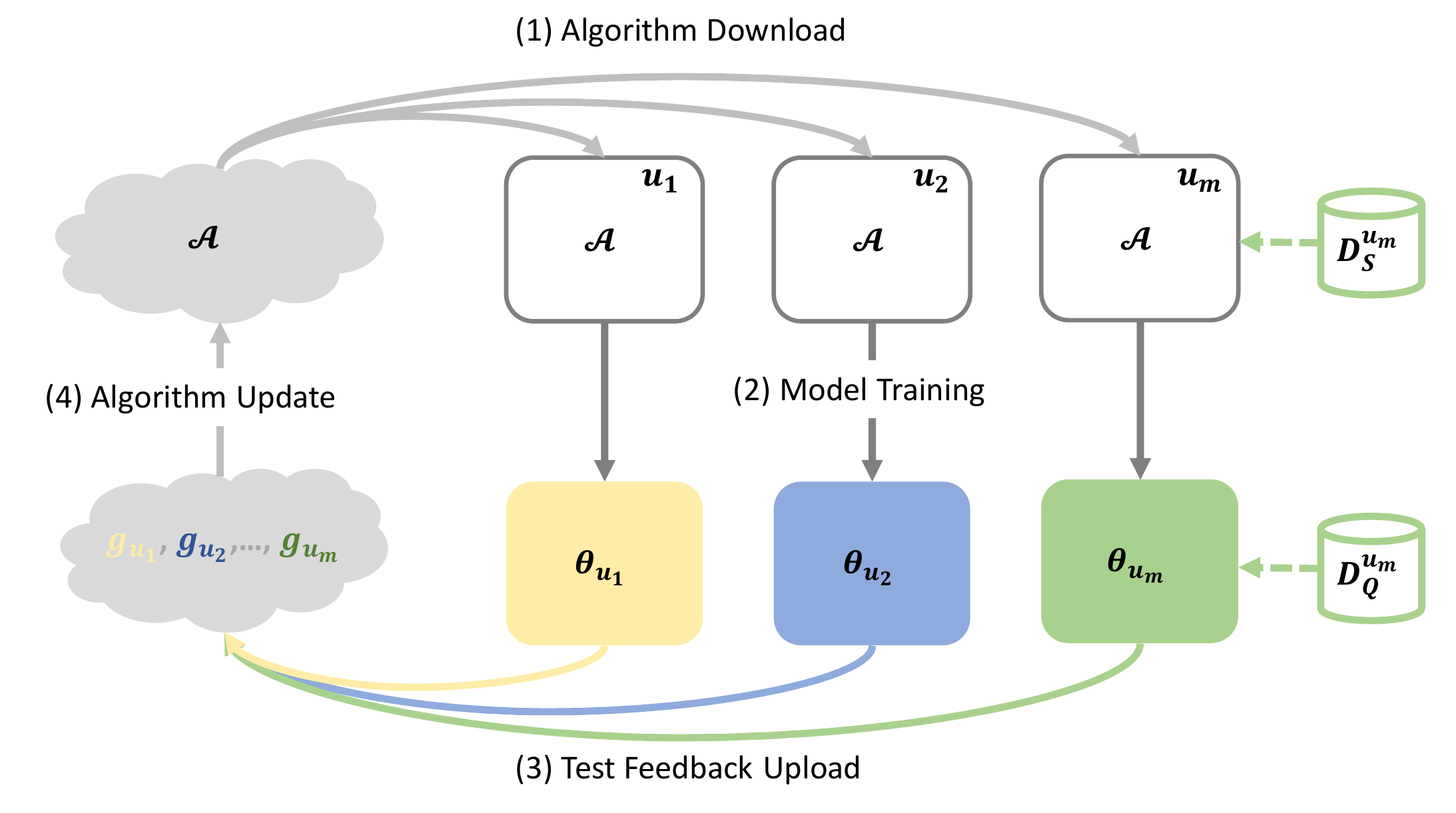}
    \caption{Workflow of the federated meta-learning framework.}
    \label{fig:workflow}
    \vspace{-15pt}
\end{figure*}

Initialization based meta-learning algorithms like MAML ~\cite{Finn2017} are well known for rapid adaptation and good generalization to new tasks, which makes it particularly well-suited for federated setting where the decentralized training data is non-IID and highly personalized. Inspired by this, we propose a federated meta-learning framework which differs significantly from prior work in federated learning. Our work bridges the meta-learning methodology and federated learning. In meta-learning, a parameterized \emph{algorithm} (or \emph{meta-learner}) is slowly learned from a large number of tasks through a meta-training process, where a specific model is fast trained by the algorithm in each task. A task typically consists of a \emph{support} set and a \emph{query} set that are disjoint from each other. A task-specific model is trained on the \emph{support} set and then tested on the \emph{query} set, and the test results are used to update the algorithm. By contrast, in federated meta-learning, an algorithm is maintained on the server, and is distributed to the clients for model training. In each episode of meta-training, a batch of sampled clients receives the parameters of the algorithm and performs model training. Test results on the \emph{query} set are then uploaded to the server for algorithm update. The workflow of our framework is illustrated in Figure \ref{fig:workflow}.

\textbf{Comparing federated meta-learning with federated learning.}
The federated meta-learning framework may seem similar to federated learning, except that the information transmitted between the server and clients is (parameters of) an algorithm instead of a global model. However, we note that meta-learning is conceptually different from distributed model training, and a shared algorithm in federated meta-learning can be applied in a more flexible way than a shared model as in federated learning.
For example, in image classification, images of $n$ categories may be distributed non-uniformly among clients, where each client possesses at most $k$ categories with $k \ll n$. Federated learning would need to train a large $n$-way classifier to utilize data from all clients, while a $k$-way classifier suffices since it makes predictions for one client each time. The large model increases communication and computation costs. It is possible to send a client only a portion of the model to update the relevant parameters, but this would require knowledge of the client's private data beforehand to decide the portion. In meta-learning, on the other hand, the algorithm can train tasks containing different categories. For instance, the Model-Agnostic Meta-Learning (MAML) algorithm~\cite{Finn2017}, to be described in details in the Federated Meta-Learning Section, could provide the initialization for a $k$-way classifier by meta-training on $k$-way tasks, regardless of the specific categories. Therefore in the federated meta-learning framework, we can use MAML to meta-train a $k$-way classifier initialization with all the $n$ categories. In this way, federated meta-learning has considerably lower communication and computation costs.

\textbf{Contributions.}
We focus on the algorithm design aspect of federated setting, for which we present a new framework alongside extensive experimental results. Our contributions are threefold. First, we show that meta-learning is a natural choice for federated setting and propose a novel federated meta-learning framework named \texttt{FedMeta} that incorporates the meta-learning algorithms with federated learning. The framework allows sharing parameterized algorithm in a more flexible manner, while preserving client privacy with no data collected onto servers.
We integrate gradient based meta-learning algorithms MAML and Meta-SGD into the framework for illustration.
Second, we run experiments on LEAF datasets to compare the running examples contained in our \texttt{FedMeta} framework with the baseline FedAvg in terms of accuracy, computation cost and communication cost. The results show that \texttt{FedMeta} achieves higher accuracies with less or comparable system overhead. Third, we apply \texttt{FedMeta} to an industrial recommendation task where each client has highly personalized records, and experimentally show that meta-learning algorithms achieve higher accuracies for recommendation tasks than federated or stand-alone recommendation approaches.

	\section{Related Work}
\label{sec:related work}

\textbf{Initialization Based Meta-Learning.} In meta-learning, the goal is to learn a model on a collection of tasks, such that it can solve new tasks with only a small number of samples~\cite{chen2019closer}. As one promising direction to meta-learning, initialization based methods has recently demonstrated effectiveness by “learning to fine-tune”. Among the various methods, some are focused on learning an optimizer such as the LSTM-based meta-learner \cite{ravi2017optimization} and the Meta Networks with an external memory \cite{munkhdalai2017meta}. Another approach aims to learn a good model initialization~\cite{Finn2017,li2017meta,nichol2018first,nichol2018reptile}, such that the model has maximal performance on a new task with limited samples after a small number of gradient descents. All of the work mentioned above only explore the setting where the tasks have a unified form (e.g., 5-way 5-shot for image classification). In this work, we fill this gap by studying meta-learning algorithms on real-world federated datasets. We focus our attention on model initialization methods where the algorithms are model- and task-agnostic and can be deployed out of the box, as the tasks and models in the federated setting vary. To the best of our knowledge, our proposed framework is the first to explore the federated setting from the meta-learning perspective.

\textbf{Federated Learning.} To handle the statistical and systematic challenges in the ferderated setting, many optimization methods have been proposed~\cite{McMahan2016,sahu2018convergence,lin2018don}, which have significantly improve the convergence performance and system overhead. For example,~\cite{McMahan2016} adds more computation to each client and~\cite{sahu2018convergence} incorporates a proximal term into local objective function which penalizes large changes from the current global model. However, all works mentioned above aim to learn a large global model across decentralized data, which increases communication and computation cost, and limits their ability to deal with non-IID data as well as heterogeneous structure among clients. In contrast, \cite{smith2017federated} considers multi-task learning in the federated setting and proposes a communication-efficient primal-dual optimization method, which learns separate but related models for each client. However, this approach does not apply to non-convex deep learning models, where strong duality is no longer guaranteed. 

Similar to \cite{smith2017federated}, the federated meta-learning framework proposed by us treats each client as a task. Instead of training a global model that ingests all tasks, we aim to train a well-initialized model that can achieve rapid adaptation to new tasks. The intuition behind meta-learning algorithms is  to extract and propagate internal transferable representations of prior tasks. As a result, they can prevent overfitting and improve generalization on new tasks, which shows the potential in handling the statistical and systematic challenges of federated setting. Besides, we consider model-agnostic meta-learning algorithms including MAML, first-order MAML (FOMAMAL) and Meta-SGD, so our framework could handle the non-convex problems well.

	\section{Federated Meta-Learning}
\label{sec:model}

In this section, we elaborate the proposed federated meta-learning framework. We first discuss the meta-learning approach and present Model-Agnostic Meta-Learning (MAML)~\cite{Finn2017} and Meta-SGD~\cite{li2017meta} algorithms as running examples. Then we describe how meta-learning algorithms are implemented in the federated setting.

\subsection{The Meta-Learning Approach}

The goal of meta-learning is to \emph{meta-train} an algorithm $\mathcal{A}$ that can quickly train the model, e.g. a deep neural network, for new tasks. The algorithm $\mathcal{A}_\varphi$ is in general parameterized, where its parameter $\varphi$ is updated in the meta-training process using a collection of \emph{tasks}.
A task $T$ in meta-training consists of a \emph{support} set $D_S^T=\{(x_i, y_i)\}_{i=1}^{|D_S^T|}$ and a \emph{query} set $D_Q^T=\{(x'_i, y'_i)\}_{i=1}^{|D_Q^T|}$, both of which contain labeled data points.
The algorithm $\mathcal{A}$ trains a model $f$ on the support set $D_S^T$ and outputs parameter $\theta_T$, which we call \emph{inner update}. The model $f_{\theta_T}$ is then evaluated on the query set $D_Q^T$, and some test loss $\mathcal{L}_{D_Q^T}(\theta_T)$ is computed to reflect the training ability of $\mathcal{A}_\varphi$. Finally,  $\mathcal{A}_\varphi$ is updated to minimize the test loss, which we call \emph{outer update}. Note that the support and query sets are disjoint to maximize the generalization ability of $\mathcal{A}_\varphi$. Meta-training proceeds in an episodic manner, where in each episode a batch of tasks are sampled from a task distribution $\mathcal{T}$ over a meta-training set. Therefore, the algorithm $\mathcal{A}_\varphi$ is optimized with the following objective:

\begin{align}
\min_{\varphi} \mathbb{E}_{T\sim\mathcal{T}} \left[\mathcal{L}_{D^T_{Q}} (\theta_T) \right]
= \min_{\varphi} \mathbb{E}_{T\sim\mathcal{T}} \left[\mathcal{L}_{D^T_{Q}} \left(\mathcal{A}_{\varphi} \left(D^T_{S}\right)\right)\right].
\label{eq:obj_general}
\end{align}

The MAML algorithm~\cite{Finn2017} is a representative gradient-based meta-learning method that trains the model with gradient update steps. The algorithm $\mathcal{A}$ for MAML is simply used to provide the initialization for the model. In particular, for each task $T$ the algorithm maintains $\varphi=\theta$ that serves as the initial value of the parameter of model $f$. Then $f_{\theta}$ is trained on the support set $D_S^T$, and $\theta$ is updated to $\theta_T$ using one (or more) gradient descent step with training loss $\mathcal{L}_{D_S^T}(\theta) := \frac{1}{|D_S^T|}\sum_{(x,y)\in D_S^T} \ell(f_{\theta}(x), y)$, where $\ell$ is the loss function, e.g. cross entropy for image classification tasks. Finally $f_{\theta_T}$ is tested on the query set $D_Q^T$ and the test loss $\mathcal{L}_{D_Q^T}(\theta_T) := \frac{1}{|D_Q^T|}\sum_{(x',y')\in D_Q^T} \ell(f_{\theta_T}(x'), y')$ is calculated. The optimization objective in Equation~\eqref{eq:obj_general} is instantiated as follows:
\begin{align}
\min_{\theta} \mathbb{E}_{T\sim\mathcal{T}} \left[\mathcal{L}_{D^T_{Q}} \left(\theta - \alpha \nabla\mathcal{L}_{D_S^T}(\theta)\right)\right],
\label{eq:obj_maml}
\end{align}
where $\alpha$ is the learning rate for the inner gradient update.

Based on MAML, Meta-SGD~\cite{li2017meta} takes a step further to learn the initialization $\theta$ and inner learning rate $\alpha$ at the same time. Note that the test loss $\mathcal{L}_{D_Q^T}(\theta_T)$ can be treated as a function of $\theta$ and $\alpha$, both of which can be updated in the outer loop using SGD by taking gradients on $\mathcal{L}_{D_Q^T}(\theta_T)$. Moreover, the learning rate $\alpha$ is a vector of the same dimension as $\theta$, such that $\alpha$ corresponds to $\theta$ coordinate-wise. The optimization objective of Meta-SGD can be written as
\begin{align}
\min_{\theta, \alpha} \mathbb{E}_{T\sim\mathcal{T}} \left[\mathcal{L}_{D^T_{Q}} \left(\theta - \alpha \circ \nabla\mathcal{L}_{D_S^T}(\theta)\right)\right].
\label{eq:obj_meta-sgd}
\end{align}

\IncMargin{1.5em}
\begin{algorithm}[t]
	\SetAlgoLined
	
	\caption{\texttt{FedMeta} with MAML and Meta-SGD}
	\label{algo:fed-ml}
	
	// Run on the server
	
	\textbf{AlgorithmUpdate:}
	
	\vspace{2pt}
	
	Initialize $\theta$ for MAML, or initialize $(\theta, \alpha)$ for Meta-SGD.
	
	\For {\emph{each episode $t=1,2,...$}}{
		
		Sample a set $U_t$ of $m$ clients, and distribute $\theta$ (for MAML) or $(\theta, \alpha)$ (for Meta-SGD) to the sampled clients.
		
		\For {\emph{each client $u\in U_t$ {in parallel}}}{
			
			Get test loss $g_u \leftarrow \text{ModelTrainingMAML}(\theta)$ or
			
			\nonl
			\hspace{51.5pt}$g_u \leftarrow \text{ModelTrainingMetaSGD}(\theta, \alpha)$
		}
		
		Update algorithm paramters $\theta \leftarrow \theta - \frac{\beta}{m} \sum_{u\in U_t} g_u$ for MAML or 
		
		\nonl
		\hspace{114pt}$(\theta, \alpha) \leftarrow (\theta, \alpha) - \frac{\beta}{m} \sum_{u\in U_t} g_u$ for Meta-SGD.
	}

	\vspace{10pt}
	
	// Run on client $u$
	
	\begin{multicols}{2}
		\textbf{ModelTrainingMAML($\theta$):}
		
		\vspace{2pt}
		Sample support set $D^u_{S}$ and query set $D^u_{Q}$
		
		$\mathcal{L}_{D_S^u}(\theta) \leftarrow \frac{1}{|D_S^u|}\sum_{(x,y)\in D_S^u} \ell(f_{\theta}(x), y)$
		
		$\theta_u \leftarrow \theta - \alpha \nabla \mathcal{L}_{D_S^u}(\theta)$
		
		$\mathcal{L}_{D_Q^u}(\theta_u) \leftarrow \frac{1}{|D_Q^u|}\sum_{(x',y')\in D_Q^u} \ell(f_{\theta_u}(x'), y')$
		
		$g_u \leftarrow \nabla_{\theta} \mathcal{L}_{D_Q^u}(\theta_u)$
		
		Return $g_u$ to server
		
		\nonl
		\textbf{ModelTrainingMetaSGD($\theta, \alpha$):}
		
		\vspace{2pt}
		\nonl
		Sample support set $D^u_{S}$ and query set $D^u_{Q}$
		
		\nonl
		$\mathcal{L}_{D_S^u}(\theta) \leftarrow \frac{1}{|D_S^u|}\sum_{(x,y)\in D_S^u} \ell(f_{\theta}(x), y)$
		
		\nonl
		$\theta_u \leftarrow \theta - \alpha \circ \nabla \mathcal{L}_{D_S^u}(\theta)$
		
		\nonl
		$\mathcal{L}_{D_Q^u}(\theta_u) \leftarrow \frac{1}{|D_Q^u|}\sum_{(x',y')\in D_Q^u} \ell(f_{\theta_u}(x'), y')$
		
		\nonl
		$g_u \leftarrow \nabla_{(\theta, \alpha)} \mathcal{L}_{D_Q^u}(\theta_u)$
		
		\nonl
		Return $g_u$ to server
	\end{multicols}
	
\end{algorithm}

\begin{table*}[t]
	\centering
	\caption{Statistics of selected datasets.}
	\label{table:federated_data}
	\begin{tabular}{c c c c c c c c}
	\toprule[1pt]
		\textbf{Dataset} & \textbf{Clients} & \textbf{Samples} & \textbf{Classes} & \multicolumn{2}{c}{\textbf{samples per client}} & \multicolumn{2}{c}{\textbf{classes per client}} \\
		\cline{5-8}
		& \multicolumn{3}{c}{} & mean & stdev & min & max \\
		\hline
		FEMNIST & 1,068 & 235,683 & 62 & 220 & 90 & 9 & 62 \\
		Shakespeare & 528 & 625,127 & 70 & 1183 & 1218 & 2 & 70 \\
		Sent140 & 3,790 & 171,809 & 2 & 45 & 28  & 1 & 2 \\
		Production Dataset & 9,369 & 6,430,120 & 2,400 & 686 & 374  & 2 & 36 \\
	\bottomrule[1pt]
	\end{tabular}
\end{table*}

\subsection{The Federated Meta-Learning Framework}
Under the setting of federated learning~\cite{McMahan2016}, the training data is distributed among a set of clients, and one aims to collaboratively train a model without collecting data onto the server. The model is distributed and trained on the clients, and the server maintains a shared model by averaging the updated models collected from the clients. In many practical applications, such as making recommendations for mobile phone users, the model is in turn used to make predictions for the same set of clients.

We incorporate meta-learning into the federated learning framework. The goal is to collaboratively meta-train an algorithm using data distributed among clients. Taking MAML as a running example, we aim to train an initialization for the model by using all clients' data together. Recall that MAML contains two levels of optimization: the inner loop to train task-specific models using the maintained initialization, and the outer loop to update the initialization with the tasks' test loss. In the federated setting, each client $u$ retrieves the initialization $\theta$ from the server, trains the model using a support set $D_S^u$ of data on device, and sends test loss $\mathcal{L}_{D_Q^u}(\theta)$ on a separate query set $D_Q^u$ to the server. The server maintains the initialization, and updates it by collecting test losses from a mini batch of clients.

The transmitted information in this process consists of the model parameter initialization (from server to clients) and test loss (from clients to server), and no data is required to be collected to the server.
For Meta-SGD, the vector $\alpha$ is also transmitted as part of the algorithm parameters and used for inner loop model training.

Algorithm~\ref{algo:fed-ml} illustrates the federated meta-learning framework \texttt{FedMeta} with MAML and Meta-SGD, where communication round corresponds to episode in meta-learning terminology. The algorithm is maintained in the AlgorithmUpdate procedure. In each round of update, the server calls ModelTrainingMAML or ModelTrainingMeta-SGD on a set of sampled clients to gather test losses.
To deploy the model on client $u$ after meta-training, the initialization $\theta$ is updated using the training set of $u$, and the obtained $\theta_u$ is used to make predictions.

	\section{Experiments}
\label{sec:exp}

In this section, we evaluate the empirical performance of \texttt{FedMeta} on different tasks, models, and real-world federated datasets. First, we conduct experiments on LEAF~\cite{caldas2018leaf} datasets --- a benchmark for federated settings and show that \texttt{FedMeta} can provide faster convergence, higher accuracy and lower system overhead compared with traditional federated learning approach. Second, we evaluate \texttt{FedMeta} in a more realistic scenario --- an industrial recommendation task, and demonstrate that \texttt{FedMeta} can keep the algorithm and the model at a smaller scale while maintaining higher capacities. The statistics of the selected datasets are summarized in Table \ref{table:federated_data}.

\subsection{Evaluation Scheme}

In all experiments, we randomly select 80\% of clients as training clients, 10\% clients as validation clients, and the remaining as testing clients, as we consider the ability to generalize to new clients as a crucial property of federated learning. For each client, the local data is divided into the support set and query set. We vary the fraction $p$ of data used as support set for each client to study how efficiently could \texttt{FedMeta} adapt to new users with limited data. We denote this setting by ``$p$ Support'' in the rest of this section.

As for traditional federated learning, we consider the Federated Averaging algorithm (FedAvg)~\cite{McMahan2016}, which is a heuristic optimization method based on averaging local Stochastic Gradient Descent (SGD) updates, and has been shown to work well empirically in the non-convex setting. For fair comparison, we also implement a meta-learning version of FedAvg, denonted by FedAvg(Meta). Different from the intuitive FedAvg, FedAvg(Meta) uses the support set of the testing clients to fine-tune the model initialization received from the server before testing it, which embodies the essence of meta-learning --- “learning to fine-tune”. During training process, both FedAvg and FedAvg(Meta) use all the data on the training clients.

As for federated meta-learning, we include three optimization oriented algorithms: MAML, the first-order approximation of MAML (denoted by FOMAML)~\cite{Finn2017} and Meta-SGD~\cite{li2017meta}, all of which are model-agnostic methods and can be readily implemented within our \texttt{FedMeta} framework. FOMAML is a simplified version of MAML where the second derivatives are omitted and is reported to have the similar performance to MAML while leading to roughly 33\% speed-up in computation cost~\cite{Finn2017}. So we additionally consider FOMAML when comparing the system overhead. More details about the implementation are provided in Appendix.

\subsection{LEAF Datasets}
\label{leaf}

\begin{table*}[t]
	\centering
	\caption{Accuracy results on LEAF Datasets. For FEMNIST, Shakespeare and Sent140, the models are trained for 2000, 400 and 400 rounds respectively.}
	\label{table:LEAF RESULT}
	
	\begin{tabular}{l|l|c|c|c}
    \toprule[1pt]
		\multicolumn{2}{l|}{} & 20\% Support & 50\% Support & 90\% Support \\
		\hline
		\multirow{4}{*}{FEMNIST}
		& FedAvg & 76.79\% $\pm$ 0.45\% & 75.44\% $\pm$ 0.73\% & 77.05\% $\pm$ 1.43\% \\
		& FedAvg(Meta) & 83.58\% $\pm$ 0.13\% & 87.84\% $\pm$ 0.11\% & 88.76\% $\pm$ 0.78\% \\
		& FedMeta(MAML) & 88.46\% $\pm$ 0.25\% & 89.77\% $\pm$ 0.08\% & 89.31\% $\pm$ 0.15\%\\
		& FedMeta(Meta-SGD) & \textbf{89.26\%} $\pm$ 0.12\% & \textbf{90.28\%}  $\pm$ 0.02\% & \textbf{89.31\%} $\pm$ 0.09\% \\
		\hline
		\multirow{4}{*}{Shakespeare}
		& FedAvg & 40.76\% $\pm$ 0.62\% & 42.01\% $\pm$ 0.43\% & 40.58\% $\pm$ 0.55\% \\
		& FedAvg(Meta) & 38.71\%  $\pm$ 0.51\% & 42.97\% $\pm$ 0.97\% & 43.48\% $\pm$ 0.64\% \\
		& FedMeta(MAML) & \textbf{46.06\%} $\pm$ 0.85\% & \textbf{46.29\%}  $\pm$ 0.84\% & \textbf{46.49\%} $\pm$ 0.77\%\\
		& FedMeta(Meta-SGD) & 44.72\% $\pm$ 0.72\% & 45.24\% $\pm$ 0.53\% & 46.25\% $\pm$ 0.63\% \\
		\hline
		\multirow{4}{*}{Sent140}
		& FedAvg & 71.53\% $\pm$ 0.18\% & 72.29\% $\pm$ 0.49\% & 73.38\% $\pm$ 0.38\% \\
		& FedAvg(Meta) & 70.10\%  $\pm$ 0.66\% & 73.88\% $\pm$ 0.06\% & 75.86\% $\pm$ 0.46\% \\
		& FedMeta(MAML) & 76.37\% $\pm$ 0.06\% & 78.63\%  $\pm$ 0.19\% & 79.53\% $\pm$ 0.25\%\\
		& FedMeta(Meta-SGD) & \textbf{77.24\%} $\pm$ 0.32\% & \textbf{79.38\%} $\pm$ 0.09\% & \textbf{80.94\%} $\pm$ 0.29\% \\
	\bottomrule[1pt]
	\end{tabular}
\end{table*}

We first explore LEAF \cite{caldas2018leaf} which is a benchmark for federated settings. LEAF consists of three datasets: (1) \emph{FEMNIST} for 62-class image classification, which serves as a more complex version of the popular MNIST dataset~\cite{lecun1998mnist}. The data is partitioned based on the writer of the digit/character. (2) \emph{Shakespeare}  for next character prediction, which is built from \emph{The Complete Works of William Shakespeare}~\cite{shakespeare2007complete}. Each speaking role in each play is considered as a different client. (3) \emph{Sentiment140}~\cite{go2009twitter} for 2-class sentiment classification, which is automatically generated by annotating tweets based on the emoticons presented in them. Each twitter user is considered as a client. We use a CNN model for FEMNIST, a stacked character-level LSTM model for Shakespeare and an LSTM classifier for Sent140. We filter inactive clients with fewer than $k$ records, which is set to be 10, 20, 25 for FEMNIST, Shakespeare, and Sent140 respectively. Full details about the datasets and the models we adopted are provided in Appendix.

\textbf{Accuracy and Convergence Comparsion.}  We study the performance of FedAvg and \texttt{FedMeta} framework on LEAF datasets. Considering the limited computation capacity on edge devices, we set the number of local epochs to be 1 for all methods. 

\begin{figure*}[t]
	\centering
	\includegraphics[width=\textwidth]{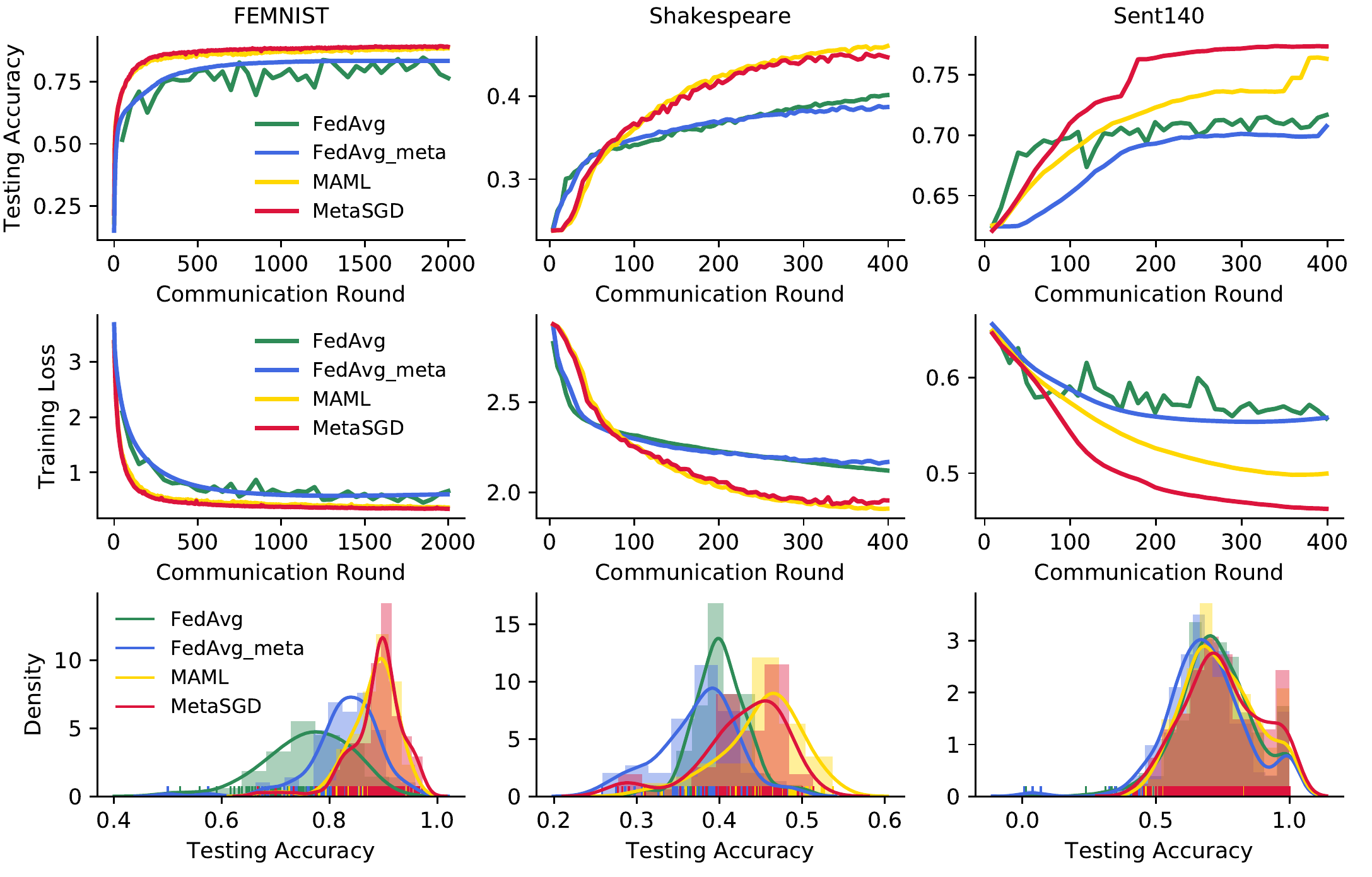}
	\caption{Performance on LEAF datasets for FedAvg and three running examples of \texttt{FedMeta}. The support fraction setting for all datasets is 20\%. Compared with intuitive FedAvg, all the running examples within \texttt{FedMeta} framework provide faster convergence and higher accuracy.}
	\label{fig:2sup}
\end{figure*}

As shown in Figure \ref{fig:2sup}, all methods within \texttt{FedMeta} framework achieve 
an increase in the final accuracy with faster and more stable convergence. We can see that MAML and Meta-SGD lead to similar convergence speed and final precision on FEMNIST and Shakespeare, while Meta-SGD performs significantly better than MAML on Sent140. 

Table \ref{table:LEAF RESULT} shows the final accuracies of four methods after several rounds of communication. First, comparing different methods, we notice that FedAvg performs significantly worse than \texttt{FedMeta}, especially on the image classification task. In contrast, MAML and Meta-SGD achieve the highest accuracies in different cases, increasing the final accuracies by 3.23\%-14.84\%. We also obeserve that FedAvg(Meta) achieves higher accuracies compared to FedAvg in most cases. However, two special cases are Shakespeare and Sent140 where the support fraction is 20\%. Unexpectedly, FedAvg(Meta) leads to a slight accuracy reduction. This may be because the model suffers an excessive deviation from the global optimal after fine-tuning on a small amount of data. Second, when we increase the support fraction $p$, the accuracies of FedAvg(Meta) and \texttt{FedMeta} rise in almost all cases. However, the growth rate of FedAvg(Meta) is larger than \texttt{FedMeta}. For example, on Shakespeare, the accuracy of FedAvg(Meta) increases by 4.77\% when the support fraction varies from 20\% to 90\%, while the accuracy of MAML increases by only 0.43\%. This shows \texttt{FedMeta} framework has better generalization ability and can efficiently adapt to new clients with limited data.

\begin{figure*}[t]
	\centering
	\begin{subfigure}[b]{0.32\textwidth}
	    \centering
        \includegraphics[width=1\textwidth]{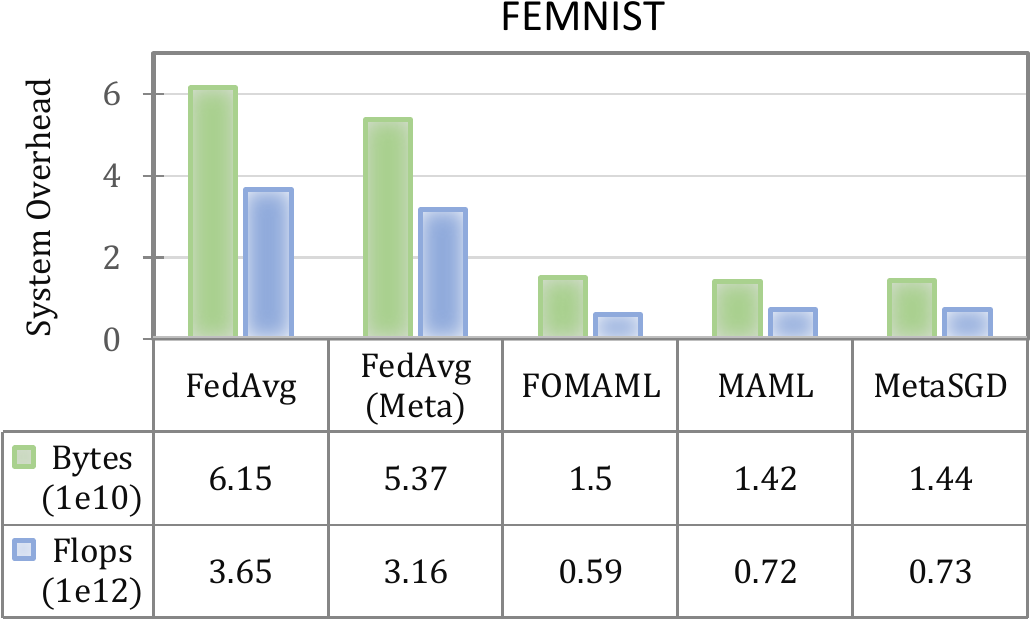}
    \end{subfigure}
    \begin{subfigure}[b]{0.32\textwidth}
        \centering
        \includegraphics[width=1\textwidth]{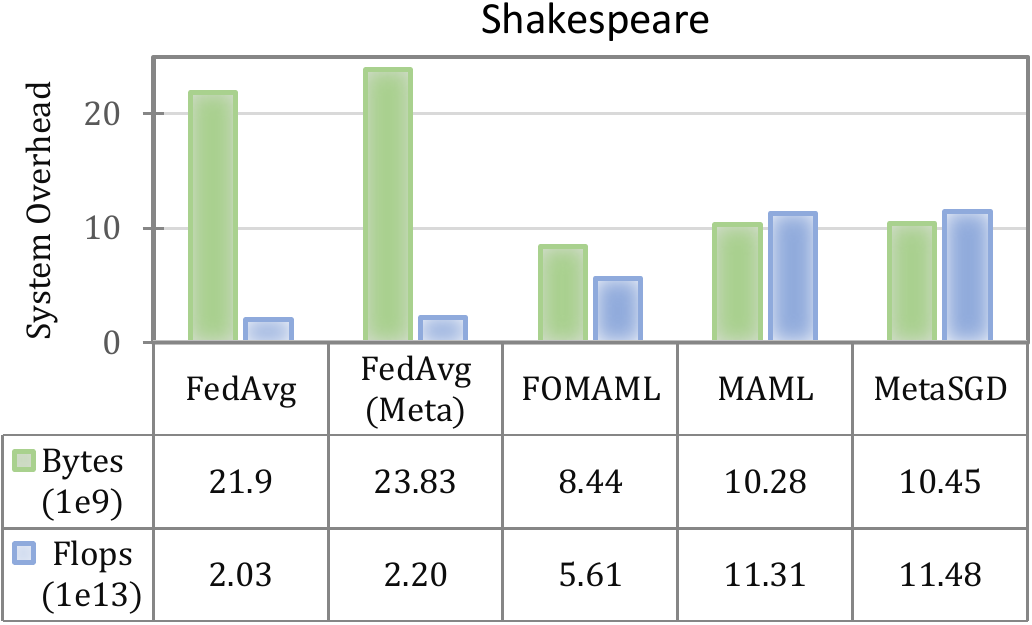}
    \end{subfigure}
    \begin{subfigure}[b]{0.32\textwidth}
        \centering
        \includegraphics[width=1\textwidth]{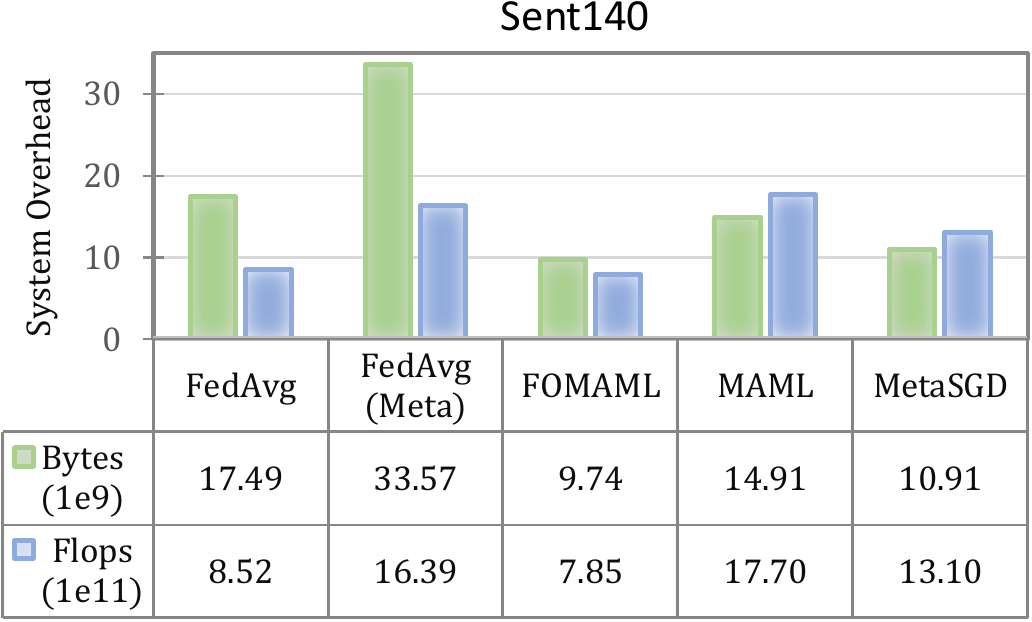}
    \end{subfigure}
	\caption{System overhead for achieving a target accuracy in different methods.The target accuracies for FEMNIST, Shakespeare and Sent140 are 74\%, 38\% and 70\% respectively.}
	\label{fig:system_overhead}
\end{figure*}

\textbf{System Overhead.}  We characterize the system budget in terms of total number of FLOPS across all devices, and total number of bytes uploaded to and downloaded from the server. Figure \ref{fig:system_overhead} shows the necessary system overhead to achieve a target test-set accuracy for different methods. Comparing communication costs, we observe that \texttt{FedMeta} achieves a reduction in required communication cost by 2.82-4.33 times in all cases. In terms of computation cost, FOMAML provides the lowest cost for FEMNIST and Sent140 due to the significant fast convergence. For shakespeare, FedAvg achieves the lowest cost, which is about 5 times less than MAML and Meta-SGD. This is because meta-learning methods produce significant computational expense by using second derivatives when back-propagating the meta-gradient. Comparing MAML and FOMAML, as expected, FOMAML reduces the computation cost for all datasets. For two language modeling tasks, FOMAML decreases the communication cost compared to MAML. But for image classification task, FOMAML increases the communication cost, which shows that the dropping of the backward pass in FOMAML has a greater effect on the convolutional network than LSTM. In general, we can flexibly choose different methods to tradeoff between communication and computation cost in the practical application.

\textbf{Fairness Comparison.} Following recent work \cite{li2019fair}, we study the fairness of FedAvg and \texttt{FedMeta} framework by comparing the final accuracy distribution averaged over multiple runs. In the last row of Figure \ref{fig:2sup}, we show the kernel density estimation of different methods. For FEMNIST, we observe that MAML and Meta-SGD not only lead to higher mean accuracies, but also achieve more centered accuracy distribution with lower variance. For Shakespeare, although \texttt{FedMeta} results in higher variance, the mean accuracy is also higher. For Sent140, the accuracy distributions around the highest peaks are almost the same for all methods. However, we see that MAML and Meta-SGD lead to more clients whose accuracies are around 100\%. Overall, \texttt{FedMeta} encourages a more fair accuracy distribution across devices for image classification task. While for language modeling tasks, \texttt{FedMeta} remains comparable fairness or sacrifices fairness for higher mean accuracy.

\subsection{Real Industrial Recommendation Task}
\label{sec:production}

\begin{table*}[t]
	\centering
	\caption{Accuracies on the production dataset. In the \mixed setting, NN-unified is trained with 80000 (gradient) steps during pre-training. In the \self setting, both LR and NN are trained locally with 100 to 10000 steps. In the \meta setting, the algorithms are locally trained with 100 steps.}
	\label{table:prod_res}
	
	\begin{tabular}{l|l|l|c|c|c|c}
	\toprule[1pt]
		\multicolumn{3}{l|}{} & \multicolumn{2}{c|}{80\% Support} & \multicolumn{2}{|c}{5\% Support} \\
		\cline{4-7}
		\multicolumn{3}{l|}{} & Top 1 & Top 4 & Top 1 & Top 4 \\
		\hline
		\mixed & \multicolumn{2}{|c|}{NN-unified (918,452 params)} & 76.72\% & 89.13\% & 66.47\% & 79,88\% \\
		\hline
		\multirow{7}{*}{\self}
		& \multicolumn{2}{|c|}{MFU} & 42.92\% & 81.49\% & 42.18\% & 72.87\% \\
		& \multicolumn{2}{|c|}{MRU} & 70.44\% & 81.43\% & 70.44\% & \textbf{81.43\%} \\
		& \multicolumn{2}{|c|}{NB} & 78.18\% & 92.57\% & 59.16\% & 72.83\% \\
		\cline{2-7}
		& \multirow{2}{*}{LR (4,160 params)}
		& 100 steps & 58.30\% & 86.52\% & 52.53\% & 75.25\% \\
		& & 10000 steps & 78.31\% & 93.70\% & 65.35\% & 77.11\% \\
	    \cline{2-7}
		& \multirow{2}{*}{NN (9,256 params)}
		& 100 steps & 57.20\% & 88.37\% & 49.89\% & 75.26\% \\
		& & 10000 steps & 83.79\% & 94.56\% & 68.87\% & 77.66\% \\
		\hline
		\multirow{4}{*}{\meta}
		& \multicolumn{2}{|l|}{MAML + LR (8,320 params)} & 47.69\% & 71.60\% & 46.75\% & 66.26\% \\
		& \multicolumn{2}{|l|}{Meta-SGD + LR (12,480 params)} & 81.70\% & 93.56\% & 72.32\% & 77.94\% \\
		& \multicolumn{2}{|l|}{MAML + NN (18,512 params)} & 83.87\% & 94.88\% & \textbf{73.08\%} & 78.02\% \\
		& \multicolumn{2}{|l|}{Meta-SGD + NN (27,768 params)} & \textbf{86.23\%} & \textbf{96.46\%} & {72.98\%} & {78.17\%} \\
    \toprule[1pt]
	\end{tabular}
\end{table*}

In order to demonstrate the effectiveness of our \texttt{FedMeta} framework on real-world application with natural partitioning of data over clients, we also evaluate \texttt{FedMeta} on a large production dataset from an industrial recommendation task. Our goal is to recommend the top-$k$ mobile services for each client based on her past records. As shown in Table \ref{table:federated_data}, there are 2400 distinct services, 9,369 clients and about 6.4 million usage records in this dataset. Besides, each client has 100 to over 5,000 records and 2 to 36 services. 

\textbf{Setting.}  We cast this recommendation task as a classification problem and consider three settings: \meta, \mixed and \self. The latter two are regarded as baselines, because we want to include some classical stand-alone recommendation algorithms for fair comparison. (1) The \meta setting corresponds to federated meta-learning approach, where we adopt a 40-class classifier instead of a 2420-class classifier adopted in the \mixed setting. Meta-learning allows for training small local models, as explained when federated meta-learning is compared with federated learning in Introduction. Two architectures are considered for this classifier: logistic regression and neural network, denoted by LR and NN respectively. (2) The \mixed type represents the federated learning approach, where a unified 2420-class classifier is first trained on the training clients and then fine-tuned to each testing client with the corresponding support set. As for the classifier, we consider a neural network with one hidden layer of 64 neurons, of which the output layer consists of 2420 neurons. We denote it by NN-unified. We avoid using deep neural networks, since we focus on studying the advantage that meta-learning algorithms could bring to training recommendation models, instead of searching for an optimal model. Moreover, in practice the model would be trained on user devices that have limited computation resources where simple models are preferable. (3) The \self setting represents the stand-alone approach where a distinct model is trained for each client with its local data. We choose the following methods for classification: most frequently used (MFU), most recently used (MRU), naive Bayes (NB) and the two architectures (LR and NN) adopted in the \meta setting. Full details about the input feature vector construction and the implementation can be found in Appendix.  

\textbf{Accuracy Comparison.}  From Table~\ref{table:prod_res}, we observe the following: (1) Comparing meta-learning algorithms in the \meta setting, with other modules and settings being the same, Meta-SGD outperforms MAML and NN outperforms LR. The simplest combination MAML + LR performs worst, implying that either the algorithm or the model should have certain complexity to guarantee performance of the \texttt{FedMeta} framework. (2) Comparing \mixed, \self and \meta, the meta-learning methods MAML + NN and Meta-SGD + NN generally outperform all baselines, though both of which has been trained with only 100 gradient steps. Another interesting observation is that MRU has the highest Top 4 accuracy in the ``5\% Support'' case. This may be because the users use a small number of services in a short period of time, and MRU is not affected by the low support fraction. However, as the support set expands, which is often the case in practice, MRU would be outperformed by meta-learning methods.

\textbf{Convergence Comparison.}  We further study the convergence performance for \meta and \self setting. The models trained in meta-learning approach achieve faster convergence compared with models trained from scratch by using (non-parametric) optimization methods, implying that training of personalized models benefit from good initialization. Full experiment results can be found in Appendix.
	
	\section{Conclusion and Future Work}
\label{sec:conclusion}

In this work, we show that meta-learning is a natural choice to handle the statistical and systematic challenges in federated learning, and propose a novel federated meta-learning framework \texttt{FedMeta}. Our empirical evaluation across a suite of federated datasets demonstrates that \texttt{FedMeta} framework achieves significant improvements in terms of accuracy, convergence speed and communication cost. We further validate the effectiveness of \texttt{FedMeta} in an industrial recommendation scenario, where \texttt{FedMeta} outperforms both stand-alone models and unified models trained by the federated learning approach.

In the future, we will explore the following directions: (1) We want to study whether the \texttt{FedMeta} framework has additional advantages in preserving user privacy from the model attack perspective~\cite{tramer2016stealing,song2017machine,szegedy2013intriguing,shokri2017membership}, as the global model shared in the current federated learning approaches still includes all users' privacy implicitly, while in \texttt{FedMeta} a meta-learner is shared. (2) We will deploy our \texttt{FedMeta} framework online for APP recommendation to evaluate its online performance, involving much engineering work yet to be done.
	
	\newpage
	\bibliographystyle{plainnat}
	\bibliography{FML}
	
	\newpage
 	\appendix
	\section{Experimental Details}
\label{appendix}

\subsection{Datasets and Models}
\label{appdataset}
\textbf{FEMNIST:} We study an 62-class image classification task on FEMNIST, which is a more complex version of the popular MNIST dataset\cite{lecun1998mnist}. The data is partitioned based on the writer of the digit/character. We consider a CNN with two 5x5 convolution layers (the first one has 32 channels, the second one has 64 channels, each of them followed with $2 \times 2$ max pooling), a fully connected layer with 2048 units and ReLU activation, and a final softmax output layer. The input of the CNN model is a flattened $28 \times 28$ image, and the output is a number between 0 and 61.

\textbf{Shakespeare:} We study a next character prediction task on Shakespeare, which is built from \emph{The Complete Works of William Shakespeare} \cite{shakespeare2007complete}. In the dataset, each speaking role in each play is considered a different client. This language modeling task can be modeled as a 53-class classification problem. We use a two layer LSTM classifier containing 256 hidden units with a 8D embedding layer. The embedding layer takes a sequence of 80 characters as input, and the output is a class label between 0 and 52.

\textbf{Sent140:} We study a 2-class sentiment classification task on Sent140, which is automatically generated by annotating tweets based on the emoticons presented in them. We use a two-layer LSTM classifier with 100 hidden units and pre-trained 300D GloVe embeddings~\cite{pennington2014glove}. The input is a senquence of 25 words, where each word is subsequently embedded into a 300 dimensional space by looking up GloVe. The output of the last dense-connected layer is a class label of 0 or 1.

\textbf{Production Dataset:} We study an industrial recommendation task on a production dataset. There are 2400 distinct services and 9369 clients, where each client has 100 to over 5000 records and 2 to 36 services. In each usage record, the label is a service that has been used by the user, and the feature contains service features (e.g., service ID and etc), user features (e.g. last used service and etc) and the context features (e.g. battery level, time, and etc). We conduct experiment with three model architectures: NN-unified, NN and LR. The details of these models are provided in the Experiment Section.

\subsection{Implementation Details}
\label{appim}
\textbf{Libraries:} We implement FedAvg, FedAvg(Meta), FedMeta with MAML, FOMAML and Meta-SGD in TensorFlow \cite{abadi2016tensorflow} which allows for automatic differentiation through the gradient update(s) during meta-learning. We use Adam \cite{kingma2014adam} as the local optimizer for all approaches. 

\textbf{Evaluation} There are two common ways to define testing accuracy in the federated setting, namely accuracy with respect to all data points and accuracy with respect to all devices. In this work, we choose the former. As for sampling schemes, we uniformly sample clients at each communication round and average the local models with weights proportional to the number of local data points when updating the algorithm on the server. For fair comparison, the division of clients and the division of support/query sets remain the same for all methods.

\textbf{Hyperparameters:} For each LEAF dataset, we tune the numbers of active clients per round. For FEMNIST, Shakepseare and Sent140, the numbers of active clients are 4, 50 and 60 respectively. We also do a grid search on the learning rates, the best ones are provided in Table \ref{table:lr}.

\begin{table}[ht]
	\centering
	\caption{Learning rates setup for LEAF experiments.}
	\label{table:lr}
	
	\begin{tabular}{l|c|c}
    \toprule[1pt]
		& FedAvg & FedMeta($\alpha$, $\beta$)\\
		\hline
		FEMNIST & 10e-5 & (0.001,0.0001)\\
		Shakespeare & 0.001 & (0.1, 0.01)\\
		Sent140 & 0.001 & (0.001, 0.0001)\\
	\bottomrule[1pt]
	\end{tabular}
\end{table}

\subsection{Additional Experiments on LEAF}
\begin{figure*}[ht]
	\centering
	\includegraphics[width=\textwidth]{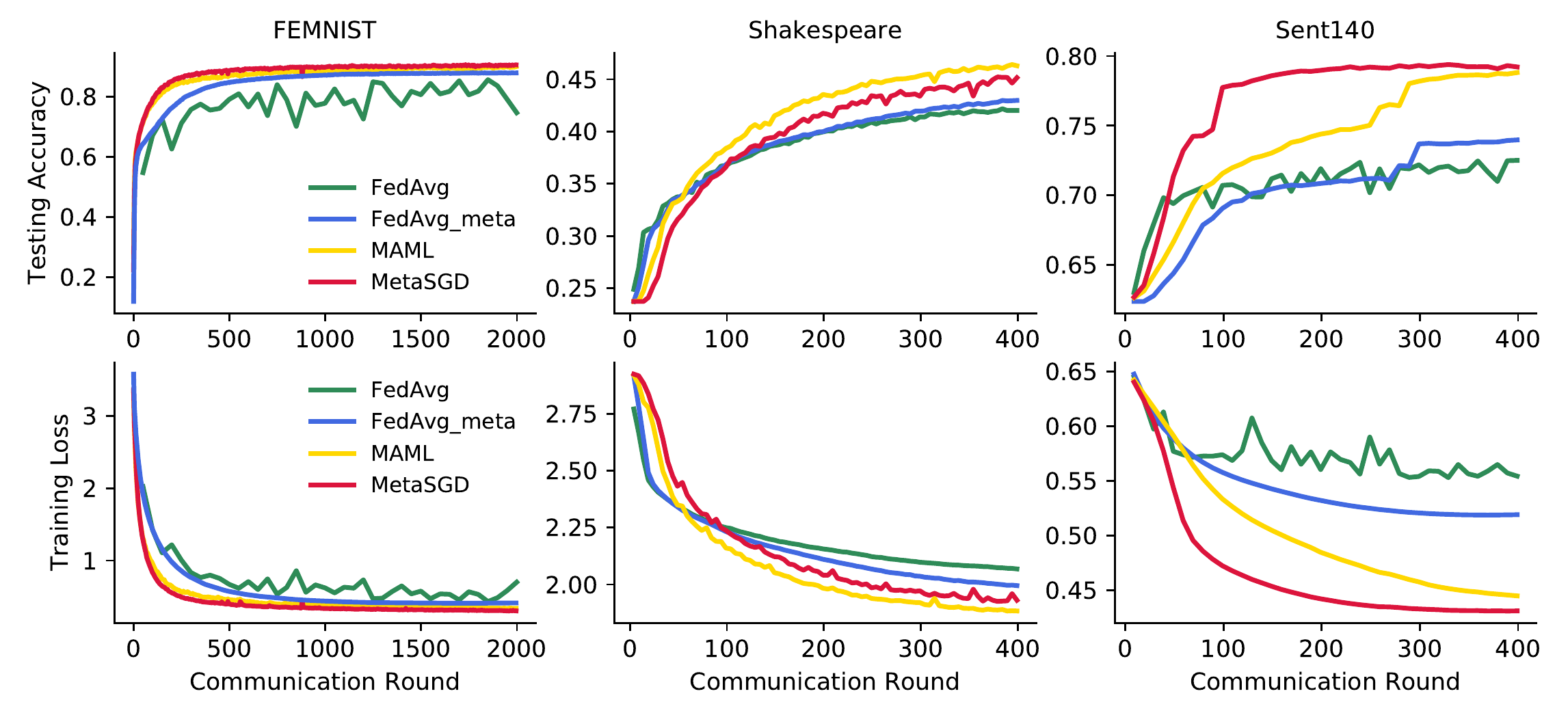}
	\caption{Performance on LEAF datasets for FedAvg and three running examples of \texttt{FedMeta}. The support fraction setting for all datasets is 50\%}
	\label{fig:5sup}
\end{figure*}

\begin{figure*}[ht]
	\centering
	\includegraphics[width=\textwidth]{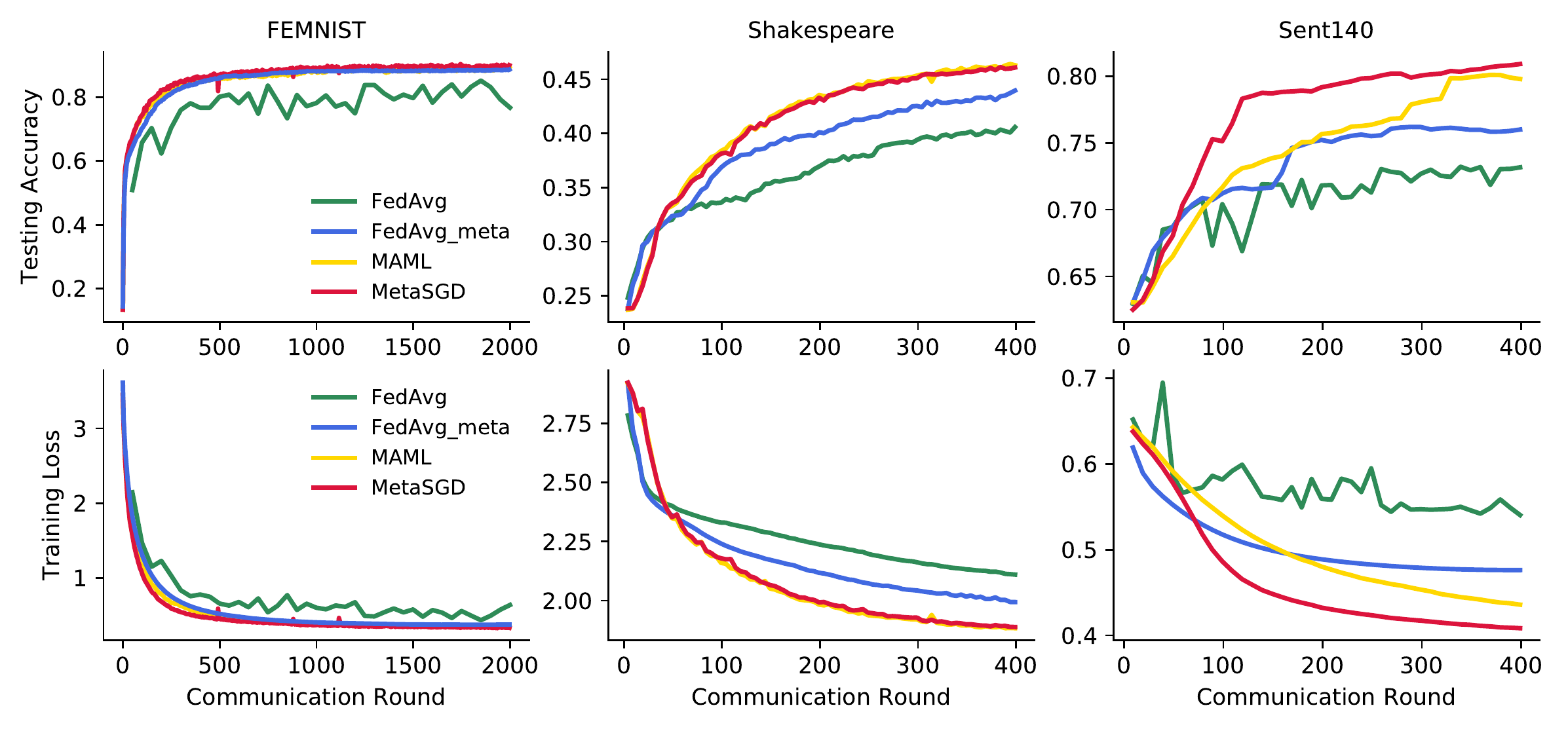}
	\caption{Performance on LEAF datasets for FedAvg and three running examples of \texttt{FedMeta}. The support fraction setting for all datasets is 90\%}
	\label{fig:9sup}
\end{figure*}

As shown in Figure \ref{fig:5sup} and Figure \ref{fig:9sup}, we provide the convergences curves when the support fractions are 50\% and 90\%. We notice that the gap between FedAvg(Meta) and \texttt{FedMeta} is shrinking when we increase the support fraction. An example is that the convergence curves of FedAvg(Meta) and \texttt{FedMeta} almost coincide when the support fraction is 90\%. This is consistent with previous industrial experience that meta-learning methods do have a greater advantage in the low-data regime than in medium- or large- data regime.

\subsection{Additional Experiments on Industrial Task}
\label{appindustry}
 
 \textbf{Input Feature Vector Construction:} 
  The input feature vectors are constructed differently for different approaches. For both \meta and \self, user-specific models are trained, and the feature vector encodes each usage record with dimension 103. For \mixed, a unified model is trained across all users. To improve the prediction accuracy, the feature vector further encodes user ID and service ID, which has dimension 11892. 

\textbf{Setting:}
 We randomly sample 7000 users as training users, some of them are used in pre-training (for baselines) or meta-training (for meta-learning methods), and the remaining ones as testing users. For each testing user, the last 20\% records in chronological order are used as query set, and for ``$p$ Support'' case, the $p$ fraction of records are used as support set. In each step in \mixed setting, a batch of 500 data points (or half of the user records, whichever is smaller) are sampled from a single user. In the \self setting, the batch size is 500 or half size of the query set. In the \meta setting, the algorithms are trained with 20000 episodes, each of them consists one task from a single training user during meta-training. In each meta-training task, 500 data points (or half of the user records) are sampled, among which the first 80\% are used as support set and the remaining as query set.

\begin{figure*}[ht]
	\centering
	\begin{subfigure}[t]{0.49\textwidth}
		\centering
		\includegraphics[width=0.8\textwidth]{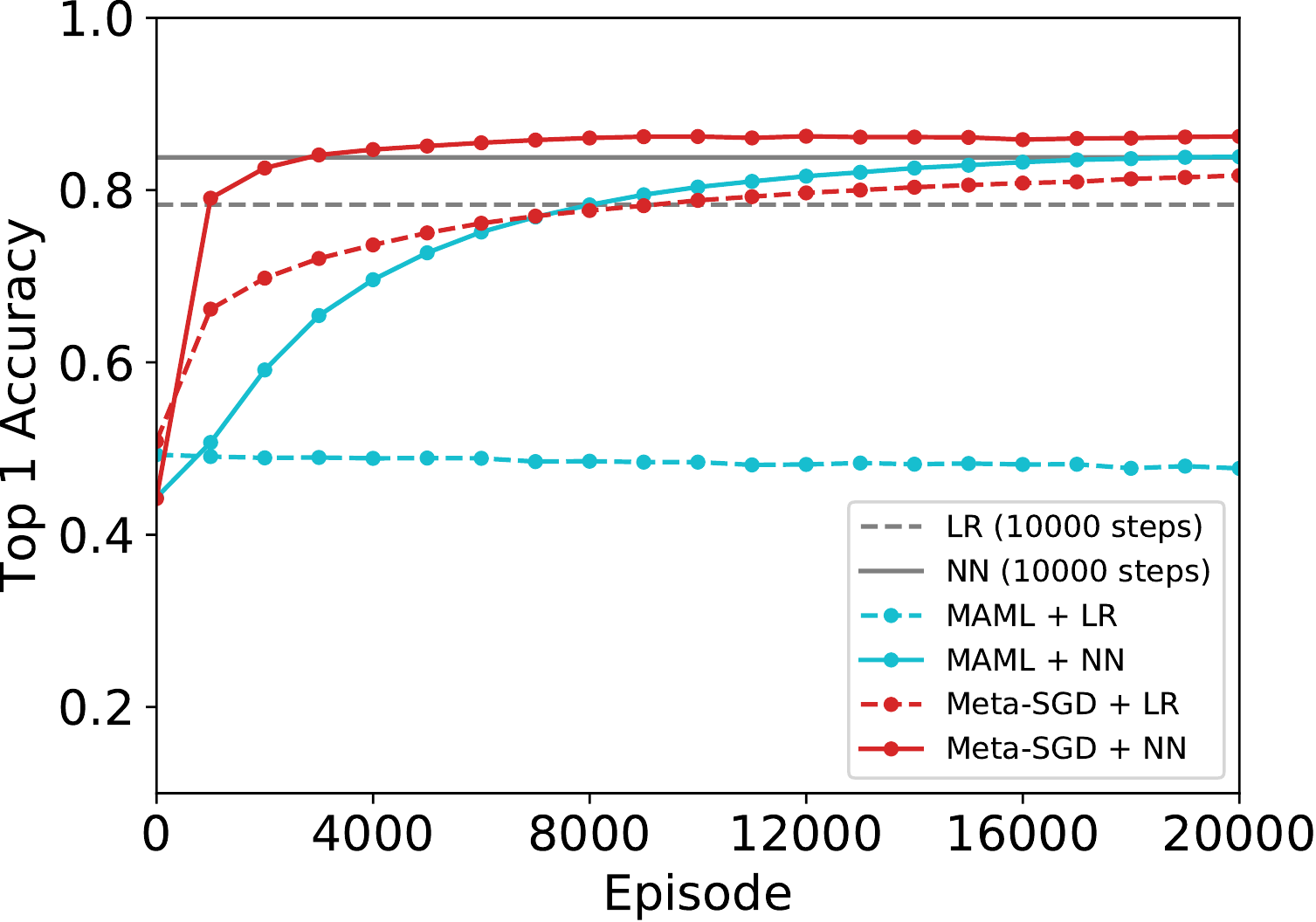}
	\end{subfigure}
	\begin{subfigure}[t]{0.49\textwidth}
		\centering
		\includegraphics[width=0.8\textwidth]{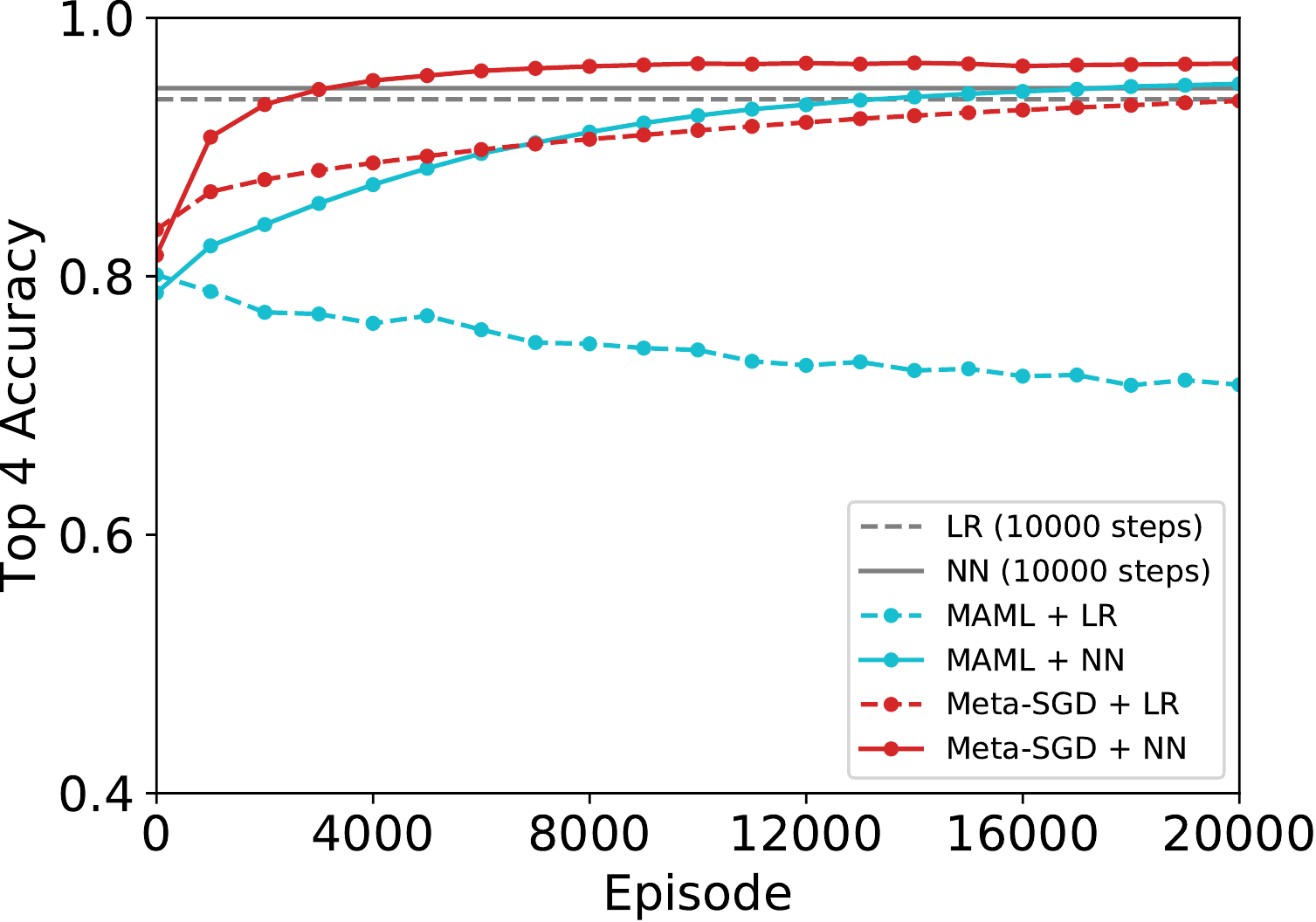}
	\end{subfigure}
	\caption{Convergence of Top 1 (left) and Top 4 (right) accuracies of meta-learning methods (``80\% Support'' case).}
	\label{fig:convergence_meta}
\end{figure*}

\textbf{Convergence Comparison:}
We further study the convergence of meta-learning methods in terms of the number of episodes during meta-training. as shown in Figure~\ref{fig:convergence_meta}. The performance of MAML + LR surprisingly becomes worse as meta-training proceeds, especially for the Top 4 recommendation. The other three methods converge within 20000 episodes, while Meta-SGD converges faster than MAML. In particular, Meta-SGD + NN already outperforms the best baseline NN after 4000 episodes.

\begin{figure*}[ht]
	\centering
	\begin{subfigure}[t]{0.49\textwidth}
		\centering
		\includegraphics[width=0.8\textwidth]{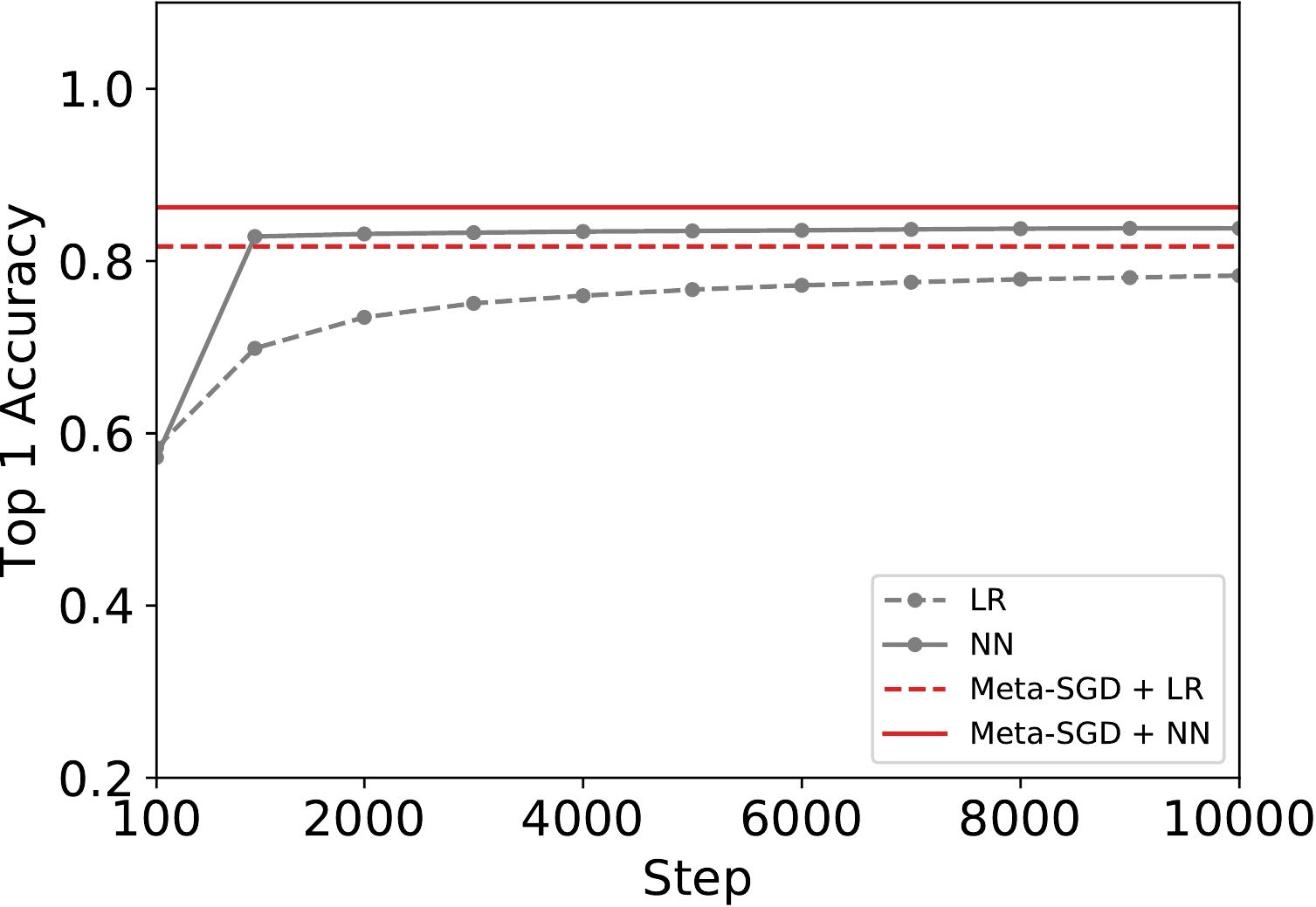}
	\end{subfigure}
	\begin{subfigure}[t]{0.49\textwidth}
		\centering
		\includegraphics[width=0.8\textwidth]{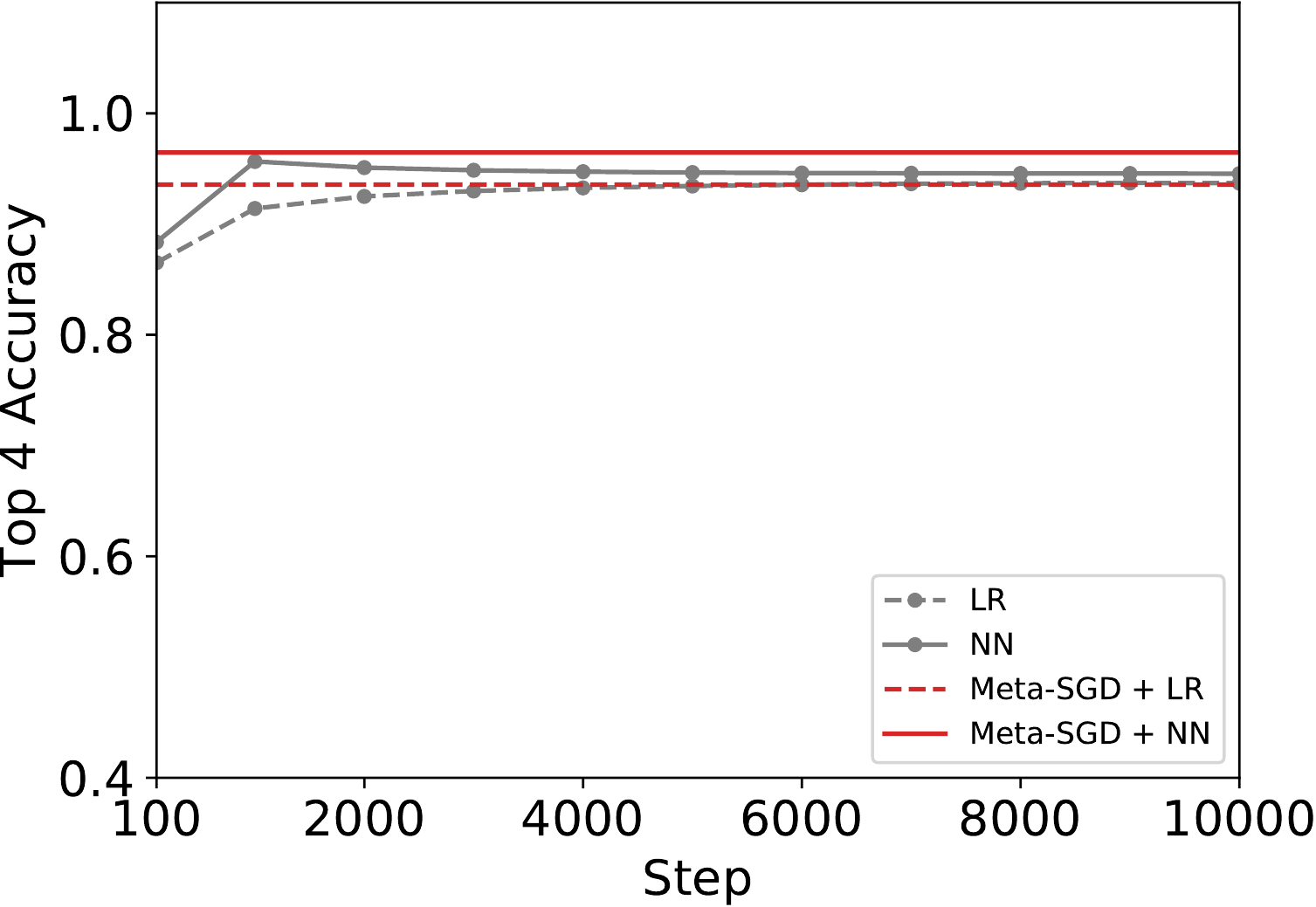}
	\end{subfigure}
	\caption{Convergence of Top 1 (left) and Top 4 (right) accuracies of \self baselines LR and NN (``80\% Support'' case).}
	\label{fig:convergence_self}
\end{figure*}

In Table~\ref{table:prod_res}, the \self baselines --- LR and NN perform significantly better when the number of training gradient steps increases from 100 to 10000. It remains to see how many steps are sufficient to train the models. Figure~\ref{fig:convergence_self} shows that LR and NN indeed converge in 10000 steps, where the accuracy of LR is below or marginally above Meta-SGD + LR, while the accuracy of NN is below Meta-SGD + NN. We stress that Meta-SGD trains the models with only 100 steps, which is much more efficient than training the models from scratch by using (non-parametric) optimization algorithms.

\end{document}